\documentclass[11pt]{article}

\usepackage{url}
\usepackage{authblk}
\usepackage{amsmath} 
\usepackage{amsfonts} 
\usepackage{bm}
\usepackage{adjustbox}
\usepackage{graphicx}
\usepackage{mdframed}
\usepackage{xcolor}

\usepackage[utf8]{inputenc}
\usepackage[greek,english]{babel}
\usepackage{lmodern}

\providecommand{\keyword}[1]
{
  \footnotesize	
  \textbf{\textit{Keywords---}} #1
}

\usepackage{color}

\makeatletter
\let\@fnsymbol\@arabic
\makeatother
\title{Multilingual Name Entity Recognition and Intent Classification Employing Deep Learning Architectures}

\author[$\dag$]{Sofia Rizou}
\author[*]{Antonia Paflioti}
\author[$\dag$]{Angelos Theofilatos}
\author[$\ddagger$]{Athena Vakali}
\author[$\dag$]{George Sarigiannidis}
\author[$\dag,\mathsection$]{Konstantinos Ch. Chatzisavvas\thanks{\noindent Corresponding author:\, k.chatzisavvas@msensis.com }}

\affil[$\dag$]{mSensis S.A., VEPE Technopolis, GR-55535, Thessaloniki, Greece}
\affil[*]{CRI S.A., 110 Pentelis Str., GR-15126, Athens, Greece}
\affil[$\ddagger$]{Informatics Department, Aristotle University of Thessaloniki, GR-54124 Thessaloniki, Greece}
\affil[$\mathsection$]{Mathematics Department, Aristotle University of Thessaloniki, GR-54124 Thessaloniki, Greece}

\title{Multilingual Name Entity Recognition and Intent Classification Employing Deep Learning Architectures}

\date{ }

\begin{document}
\maketitle


\begin{abstract}
\noindent Named Entity Recognition and Intent Classification are among the most important subfields of the field of Natural Language Processing. Recent research has lead to the development of faster, more sophisticated and efficient models to tackle the problems posed by those two tasks. In this work we explore the effectiveness of two separate families of Deep Learning networks for those tasks: Bidirectional Long Short-Term networks and Transformer-based networks. The models were trained and tested on the ATIS benchmark dataset for both English and Greek languages. The purpose of this paper is to present a comparative study of the two groups of networks for both languages and showcase the results of our experiments. The models, being the current state-of-the-art, yielded impressive results and achieved high performance.
\end{abstract}

\noindent \keyword{Named Entity Recognition, Intent Classification, Natural Language Understanding, Deep Learning, LSTM networks, Transformer networks, Conversational Agents}

\bigskip

\begin{footnotesize} \noindent{Published in \emph{Simulation Modelling Practice and Theory}, Vol.~120, 102620 (2022)} \end{footnotesize} \\
\begin{footnotesize}\noindent \url{https://doi.org/10.1016/j.simpat.2022.102620}\end{footnotesize}



\section{Introduction}\label{sec:intro}

We are living in an era where messaging applications are strongly linked to all sorts of our daily activities, and in fact 
these applications have already overtaken social networks as can be indicated in the BI Intelligence 
Report \cite{biint2015}. 
The consumption of messaging platforms is further expected to grow significantly 
in the coming years; hence this is a huge opportunity for different businesses to gain insight on what people 
are actively engaged with. 
In this age of instant gratification, consumers expect companies to respond  
quickly or with minimum delay and this, of course, requires a lot of time and effort for the company to invest 
in their workforce. 
Thus, it’s now the right time for any organization to think of new ways to stay connected with 
the end-user. 
Many organizations undergoing a digital transformation have already started harnessing the power 
of Artificial Intelligence (AI) in the form of AI-assisted customer support systems, talent screening using AI-assisted interviews, etc. 
There are even numerous conversational AI applications including Siri, Google Assistant, personal travel assistants, 
all of which provide personalized user experience. 
 


Natural Language Processing (NLP) is a subfield of AI that assists computers with 
understanding, interpreting and manipulating human language, namely speech and text. 
The constant need for modelling and simulation in NLP has been the driving force for the development of various intricate and complex networks. Simulation in NLP is simulation of human behaviour. The implementation of Deep Learning (DL) techniques in NLP, which are attributed to the many benefits of this area, has seen a significant rise in recent years. DL is also prominent for the Named Entity Recognition (NER) and Intent Classification (IC) tasks, also known as Slot Filling (SF) and Intent Extraction (IE) respectively, that we explore in this article. All the architectures that were implemented during our experiments are DL networks which have been designed to handle sequential data. The effectiveness of DL networks in the field of NLP is largely attributed to the abundance of computational resources which are now more readily available than ever before. NER and IC are two of the most important tasks of NLP and they have gained popularity in various applications, which is why an efficient and practical NER and IC system has substantial commercial value. 


A prime example is the popularity of intelligent chatbots, aiming to reform 
costumer service by enhancing customer experience and offering companies new opportunities to simultaneously 
improve the customers’ engagement process and the operational efficiency by reducing the typical cost of customer service. 
 
NLP applications utilize a variety of methods to approach various tasks, such as the ones mentioned above. 
For these tasks to be successful, the machine must perform contextual extraction, which essentially 
is the automated retrieval of structured information from text-based sources and it is the main problem 
that needs solving for a conversational agent to be helpful. 
For conversational agents in particular, contextual extraction consists of two very important sub-tasks: NER and IC. 
IC is a type of NLU 
task that helps to understand the type of action conveyed in the sentence and all its participating parts -- 
essentially it captures the general meaning of the sentence. 
NER is the sub-task of information extraction that aims to locate and classify named entities found 
in unstructured text into pre-defined categories, such as person names, locations, organizations, monetary values, etc 
and serves as the foundation for most natural language applications. An entity can be any word or series of 
words that refer to the same thing consistently. 
Formally, given a sequence of tokens $s = [x_{1}, x_{2},\ldots, x_{N}]$, NER outputs a list of tuples 
$[I_{s}, I_{e}, t]$ where  $I_{s}\in [1,n]$ indicates the start of the entity inside the sequence, $I_{e}\in [1,n]$ is the respective 
end of the entity and $t$ is the selected label from the category set.  
 
Early NER systems noted huge success in achieving good performance with the cost of human engineering 
in designing domain-specific features and rules. 
In recent years DL empowered by continuous real-valued vector representations and semantic 
composition through nonlinear processing, has been employed in NER systems, yielding state-of-the-art performance. 
The last few years have seen a revolution in the way NLP is implemented and innovated thanks to the (re)introduction 
of neural networks and DL. The key improvements over pre-neural approaches are undoubtedly 
the considerable reduction of data sparsity and the compactness of the lexical representations. 
These come, however, at the cost of flattening information and, at least initially, conflating the meanings of ambiguous 
words into a single vector representation. 
More importantly, the biggest challenge of neural approaches is their accountability in the future, 
i.e., the ability to explain their outputs in a way that makes it possible to apply remedies. 
While this is an obvious issue for driverless cars, it is also important that an intelligent system should be able 
to explain the process followed for understanding text, especially if a decision has to be taken 
(e.g., in booking a restaurant or fixing an appointment by interacting with a vocal assistant).

In this paper we examine several state-of-the-art architectures which largely solve the NER and IC tasks 
and we compare the results we received from the experiments we conducted for this aim. 
For the first experiments, the NER and IC tasks were addressed and solved separately, and 
the models used for these tasks were evaluated both independently of each other but also as a unified model. 
For the NER task, a Bidirectional Long-Short Term Memory (Bi-LSTM) network serves as the basis of the network 
and we explore variations of the model by testing different types of word and character embeddings. 
For the independent IC task, a Support-Vector Machine (SVM) architecture is utilized. 
For the second series of experiments we conducted, the above tasks are approached with the notion 
that they are highly dependent of each other, meaning that the named entities found in a sentence 
are strongly connected to the intent of the sentence and that the intent is largely defined by the entities that comprise the sentence. 
The architectures we propose for this joint model are Transformer-based \cite{vaswani2017attention} and, 
more specifically, Bidirectional Encoder Representations from Transformers, commonly known as BERT  \cite{devlin2018bert}. 
All of the proposed models were trained, tested and evaluated on the popular ATIS (Airline Travel Information Systems) dataset, 
which is considered a benchmark dataset for NER and IC. 
Our methodology might be useful for multilingual applications thus we considered the Greek translation of the ATIS dataset and assessed all of the models’ performance 
on both Greek and English, using appropriate embeddings for each language. 
The languages under inspection are English and Greek. They are fundamentally different regarding their inflection and morphology; modern English is a typical paradigm of a weak inflection language (e.g. Swedish, Danish) while modern Greek is a typical high inflection language (e.g. German, Spanish). 
As expected, with the models being the current state-of-the-art, we were able to yield superior results in comparison 
to older and more outdated networks. 
Nevertheless, it is vital to note that we do not claim this article to be exhaustive of all NER and IC works on either of the two languages.  
 

In Section~\ref{sec:relwork} we present work related to our study, In sections~\ref{sec:lstm} and~\ref{sec:metho}
we present the DL architectures employed for  NER and IC tasks, 
and the methodology of our study. A detailed description on the language models that were used to vectorize 
the data sets, both in Greek and English, is also included. 
In Section~\ref{sec:expres} we present the details of our experiments; data processing, evaluation metrics, hyperparameters and 
experimental results. 
Finally, Section~\ref{sec:conc} contains the conclusions of this comparative study, highlights and future work. 


\section{Related Work}\label{sec:relwork}

The interdisciplinary scientific field of NLP that has been active since the 1950s and was revolutionized in the late 1980s 
through the introduction of machine learning methods in language processing and further in the 2010s with the massive 
application of DL algorithms and representation models. Along with computer vision-image processing, 
it is one of the most celebrated AI applications with numerous paradigms in both theory and commercial 
applications such as machine translation \cite{wu2016google,bahdanau2014neural}, speech recognition, 
sentiment analysis and opinion mining \cite{zhao2021bert,chen2017recurrent,giatsoglou2017sentiment},
automatic text summarization \cite{maybury1999advances}, text classification \cite{indurkhya2010handbook,sammut2011encyclopedia,Goodfellow-et-al-2016}, 
question answering systems and conversational agents (chatbots) \cite{soares2020literature,bavaresco2020conversational}, etc. 
The recent sharp increase in computing speed and capacities has led to new and highly intelligent software systems, 
which are becoming progressively able to bridge the gap between human communication and computer understanding 
and supplant or augment human services.

Previous work on ATIS dataset have employed models based on Recurrent Neural Networks (RNNs) \cite{yao2013recurrent,liu2015recurrent} or models based on LSTMs \cite{yao2014spoken,kurata2016leveraging}. Although sequence level optimization was not addressed specifically, these RNN models still achieved reasonably good metrics. The incorporation of CRFs \cite{mesnil2014using,yao2014recurrent} was done to take advantage of the conditional dependencies between labels. Past and future input features were used with the introduction of bidirectional LSTMs \cite{huang2015bidirectional, graves2005bidirectional}. The correlation between intent and entities on ATIS dataset is being explored in joint architectures employing Gated Recurrent Units (GRUs) \cite{ma2016end}, attention mechanism \cite{liu2016attention, li2018self} or character embeddings to represent the semantic relations of characters \cite{daha2019deep}. Using attention mechanisms on top of LSTMs have shown empirical success for sequence encoders for sentence representations \cite{conneau2017supervised} and decoders for neural machine translation \cite{hu2020group}.

Due to their success, transformer-based models attract lots of interest from academic and industry researchers and have become the go-to architecture in NLP. Transformer was initially designed for machine translation tasks and now it is being used in most of the NLP tasks, due to its great efficiency. Multiple adaptations on existing transformer models or transformer-based models are being developed for a variety of NLP problems such as NER and IC. 
TENER \cite{yan2019tener} is an adaptive transformer model which uses the transformer encoder with a CRF layer to model not only word-level features but character-level features as well. Several architectures perform NER and IC as a combined task, such as \cite{hardalov2020enriched} which is based on a pre-trained transformer model (BERT). It introduces a pooling attention layer from the IC task to model the relationship between the two tasks for each token and adopts a global concatenated attention mechanism. In \cite{ khan2020mt} authors utilize transformers for biomedical purposes specifically. It is a flexible model which performs NER in multiple medical-context datasets. It combines pre-trained language transformer models and transfer learning. The input sentence is represented by a sequence of embedding vectors and passes through the transformer encoder for contextual representation. A shallow layer is then used to generate representations to each dataset accordingly. In \cite{ arkhipov2019tuning} authors introduce a multilingual model, trained to perform NER on four languages, Russian, Bulgarian, Czech and Polish. It proves that BERT can also be very efficient at performing specific tasks like NER in multiple languages. Document-level representation for NER is demonstrated in \cite{ schweter2020flert}, which explores further the capabilities of transformer models. By applying a simple linear layer for word-level predictions and fine-tune the model or by using transformer features in an LSTM-CRF model, NER in a document can be achieved.


\section{Deep Learning models}\label{sec:lstm}

\subsection{LSTM}

Long Short-Term Memory (LSTM) networks are improved versions of Recurrent Neural Networks (RNNs) aiming to capture long-range dependencies in sequential data.
This is due to their ability to handle backpropagated errors for a large number of steps through the use of gates that regulate the flow of information 
in the network and thus avoid the catastrophic phenomenon of exploding (vanishing) gradients that is often the case for typical RNNs. 

\begin{align}
	\begin{pmatrix} \bm{i}_{t} \\ \bm{f}_{t} \\ \bm{o}_{t} \\ \bm{g}_{t} \end{pmatrix}
	&=\begin{pmatrix} \bm{\sigma} \\ \bm{\sigma} \\ \bm{\sigma} \\ \bm{\tanh{}} \end{pmatrix}
	\bm{W}
	\begin{pmatrix} \bm{h}_{t-1} \\ \bm{x}_{t} \\ 1 \end{pmatrix}
	\\
	\bm{c}_{t}&=\bm{f}_{t}\odot \bm{c}_{t-1}+\bm{i}_{t}\odot\bm{g}_{t} 
	\\
	\bm{h}_{t}&=\bm{o}_{t}\odot \bm{\tanh{c}}_{t}	
\end{align} 
where $\bm{W}$ is the weight matrix and with $\odot$ we denote the element-wise multiplication. 

The cell state vector $\bm{c}_{t}$ carries information of the sequence (e.g., singular/plural form in a sentence). The forget gate 
$\bm{f}_{t}$ determines how much the values of $\bm{c}_{t-1}$ are kept for time $t$, the input gate $\bm{i}_{t}$ controls the amount 
of update to the cell state, and the output gate $\bm{o}_{t}$ gives how much $\bm{c}_{t}$ reveals to $\bm{h}_{t}$. Ideally, 
the elements of these gates have nearly binary values. For example, an element of $\bm{f}_{t}$ being close to $1$ may 
suggest the presence of a feature in the sequence data. Similar to the skip connections in residual networks, the cell state $\bm{c}_{t}$ 
has an additive recursive formula, which helps back-propagation and thus captures long-range dependencies.

The notion of the \emph{gate} is essential in LSTMs. Each gate's vector has neurons with values between $0$ and $1$ and they control 
the flow of information through the network. There are three different types of gates in a typical LSTM network, i.e.
\begin{itemize}
	\item \emph{Forget gate} $\bm{f}$. Controls the feedback of the cell state vector $\bm{c}_{t}$; 
	decides which information will be preserved and which 
	will be discarded
	\item \emph{Input gate} $\bm{i}$. Controls the flow of input signal in the model 
	\item \emph{Output gate} $\bm{o}$. Controls the flow of cell state vector $\bm{c}_{t}$ towards the exit. 
\end{itemize}

\subsubsection{Bidirectional LSTM}

For a given sentence of $n$ words $(x_{1},x_{2},\ldots,x_{n})$ where each word $x_{i}$ is represented as a $d$-dimensional 
vector, an LSTM computes a representation $h_{t\to}$ of the left-side context of the sentence at every word $t$. Naturally, 
generating a representation of the right-side context $h_{t\leftarrow}$ as well should add useful information. This can be achieved 
with the application of a second LSTM that reads the same sequence in reverse. The former is the \emph{forward} LSTM 
and the latter is the \emph{backward} LSTM. These are two distinct networks with different parameters. This forward and 
backward LSTM pair is referred to as a bidirectional LSTM \cite{graves2005bidirectional}. The representation of a word with 
the use of this model 
is obtained by the concatenation of its left-side and right-side context representations, $h_{t} = [h_{t\to}; h_{t\leftarrow}]$. 
Thus, a complete representation 
of a word in context is formed effectively, which is useful for numerous tagging applications.

\subsubsection{CRF Tagging}

In tasks such as NER there are strong dependencies across output labels since the “grammar” that characterizes 
interpretable sequences of tags imposes several hard constraints (e.g., I-PER cannot follow B-LOC) that would be 
impossible to model with independence assumptions. Therefore, instead of modelling tagging decisions independently, 
we model them jointly using a Conditional Random Field (CRF) \cite{lafferty2001conditional}. 
For an input sentence $\bm{X} = (x_{1},x_{2},\ldots,x_{n})$, 
we consider $P$ to be the matrix of scores output by the bidirectional LSTM network. The size of matrix $P$ is $n \times k$, 
where $k$ is the number of distinct tags, and $P_{i,j}$ corresponds to the score of the $j$th tag of the $i$th word in a sentence. 
For a sequence of predictions  $\bm{y} = (y_{1},y_{2},\ldots,y_{n})$ we define its score to be
\begin{align}
	s(\bm{X},\bm{y})=\sum_{i=0}^{n} A_{y_{i},y_{i+1}} + \sum_{i=1}^{n} P_{i,y_{i}}
\end{align}
where $A$ is the matrix of transition scores and $A_{i,j}$ represents the score of a transition from the tag $i$ to the tag $j$. 
Therefore $A$ is a square matrix of size $k$ (the number of distinct tags). 
The application of a \emph{softmax} over all possible tag sequences yields 
a probability for the sequence $\bm{y}$
\begin{align}
	p(\bm{y}\mid \bm{X})=
	\dfrac{\text{e}^{s(\bm{X},\bm{y})}}{\sum_{\bm{\tilde{y}}\in \bm{Y}_{\bm{X}}} \text{e}^{s(\bm{X},\bm{\tilde{y}})} } 
\end{align}
where $\bm{Y}_{\bm{X}}$ represents all possible tag sequences. 


\subsection{Transformer}\label{sec:trans}

The Transformer \cite{vaswani2017attention} in NLP is a novel architecture which aims to solve sequence-to-sequence tasks 
while handling long range dependencies with ease. It adopts the mechanism of \emph{attention}, differentially 
weighing the significance of each part of the input data.  
Similarly to RNNs, Transfomer models are designed to handle sequential input data, 
such as natural language, for tasks like translation and text summarization. However, unlike RNNs, Transfomer models 
do not necessarily need to process the data in order. 
Owing to the \emph{attention mechanism} providing context for any position of the input sequence, the Transformer 
does not need to process the beginning of a sentence before the end, as it identifies the context that confers 
meaning to every word in the sentence. The \emph{attention mechanism}, which allows for accessing all previous states of a sequence and weighing them according to a learn measure of relevancy, 
thus providing crucial information about far away tokens. Just like earlier models, the Transformer implements an encoder-decoder architecture structure.

\subsubsection{The Transformer Encoder}
The Transformer Encoder maps an input sequence of symbol representations $(x_{1},\ldots, x_{n})$ to a sequence of continuous 
representations $\bm{z} = (z_{1},\ldots., z_{n})$. Given $\bm{z}$, the decoder generates an output sequence $(y_{1},\ldots, y_{m})$ 
of symbols, one element at a time. For every step, except the first one, the model consumes the previously generated 
symbols as additional input when generated the next.  


The Encoder is composed by a stack of 6 identical layers, each of which has two sub-layers. 
The first is a multi-head self-attention mechanism, and the second is a simple, position-wise fully connected feed-forward network. 
A residual connection around each of the two sub-layers is employed, followed by layer normalization.

\subsubsection{The Transformer Decoder} 
The Transformer Decoder is also composed of 6 identical layers. Aside from the two sub-layers, the decoder inserts a third sub-layer, 
which performs multi-head attention over the output of the encoder stack. 
Similarly to the encoder, residual connections around each of the sub-layers are implemented, followed by layer normalization.  
The attention function maps a query and a set of key-value pairs to an output, where the query, keys, values and output are all vectors. 
The output is computed as a weighted sum of the values and the weights of the individual values are calculated by a compatibility 
function of the query with the respective key.  
The attention mechanism implemented in the Transformer architecture is called “Scaled-Dot Product Attention”. 
The input consists of queries and keys of dimensions $d_{k}$ and values of dimensions $d_{v}$. 
The dot product of the query with all the keys is computed and then divided by $\sqrt{d_{k}}$, and a softmax function is applied 
in order to calculate the weights of the values. 
Practically, attention function on a set of queries is computed simultaneously, packed together into the matrix $\bm{Q}$. 
The keys and values are also packed into matrices and the output matrix is

\begin{equation}
	\text{Attention}(Q,K,V)=\text{softmax}\left( \dfrac{QK^{T}}{\sqrt{d_{k}}} \right) V
\end{equation}

Of course, the weight matrix is
\begin{equation}
	a=\text{softmax}\left( \dfrac{QK^{T}}{\sqrt{d_{k}}} \right)
\end{equation}	


\section{Methodology}\label{sec:metho}


We implement two state-of-the-art and popular families of deep learning models, the LSTM and the Transformer. LSTM models approach and effectively tackle the NER task. We conclude our LSTM experiments with a unified model, which succeeds in simultaneously performing NER and IC by aiming to minimize the sum of the two losses for these particular tasks. The Transformer models presented in this article implement the notion that the two tasks are dependent and introduce conditional probabilities to model this dependency.

\subsection{LSTM}

In the case of LSTM family we employ the Bidirectional Long Short-Term Memory (Bi-LSTM) network as the base neural network for our experiments. 
Three models are being developed for a two-level comparison in NER. Then, two additional models, 
a Bi-LSTM--CRF with CNN character embeddings and a Bi-LSTM--CRF with LSTM character embeddings are applied in order 
to capture a deeper semantic meaning of individual words through the incorporation of features at character-level. 
With this approach we aim to compare if character-level features contribute positively to the NER task. 
Furthermore, we compare the impact of different state-of-the-art pre-trained word embedding initializations. 
In all models we use the negative \emph{log-likelihood} function as the loss function.

\subsubsection{Named Entity Recognition Model}

\begin{figure}[h!]
	\centering
	\includegraphics[width=\textwidth]{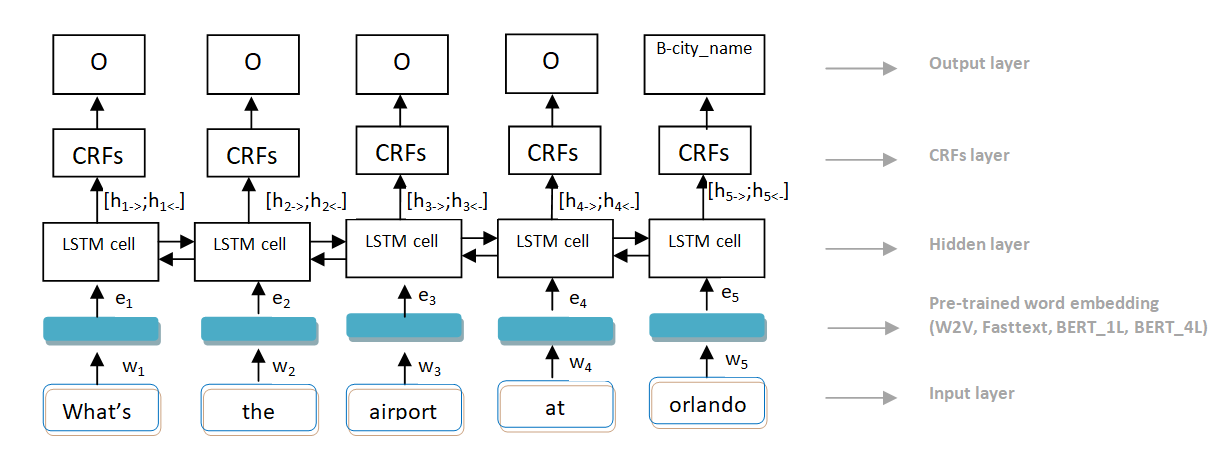}
	\caption{Bi-LSTM--CRF model}\label{fig:bilstm_crf}
\end{figure}

\noindent\textbf{Bi-LSTM--CRF}
In this model, presented Figure~\ref{fig:bilstm_crf}, we use a bidirectional RNN with long short-term 
memory units to transform word features into named entity tag scores with CRF \cite{lample2016neural}. 
We initialize the embedding layer with 
pre-trained word embeddings from three different models (Word2vec, fastText, BERT).  
Then each word embedding of the sentence passes through a Bi-LSTM unit. 
The input to CRFs layer is the concatenated forward and back hidden state of the word.

\bigskip 

\begin{figure}[h!]
	\centering
	\includegraphics[scale=0.53]{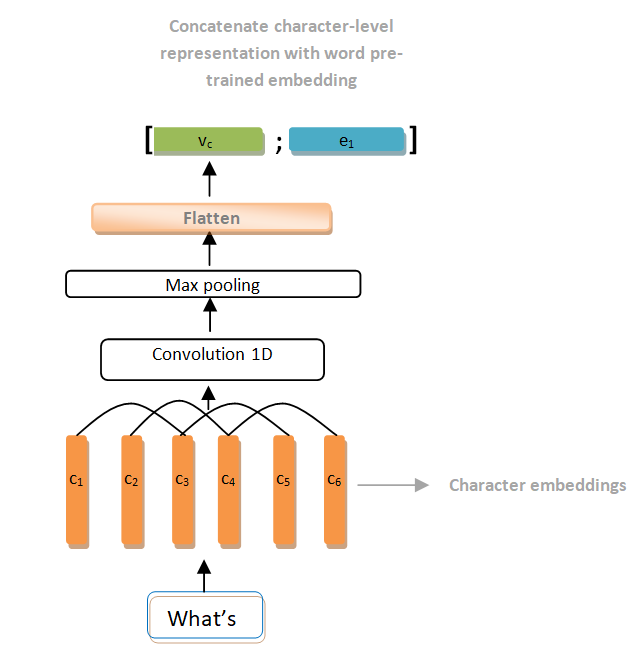}
	\caption{Character embeddings with CNN}\label{fig:bilstm_crf_cnn}
\end{figure}

\noindent\textbf{Bi-LSTM--CRF with CNN character embeddings}
The CNN character embeddings process in this model is similar to the one introduced by Ma and Hovy \cite{ma2016end}. 
For each word we employ a convolution over the characters embeddings of size $3$ and then a max pooling layer 
extract a new feature vector from the per character feature vectors (the character embeddings are initialized randomly) 
(Figure~\ref{fig:bilstm_crf_cnn}). All character feature vectors of the corresponding word are then flattened to represent 
the word from its characters perspective. The resulting $v_{c}$ vector is then concatenated with the pre-trained 
word embedding before it passes through the Bi-LSTM unit as described above.

\bigskip

\begin{figure}[h!]
	\centering
	\includegraphics[width=\textwidth]{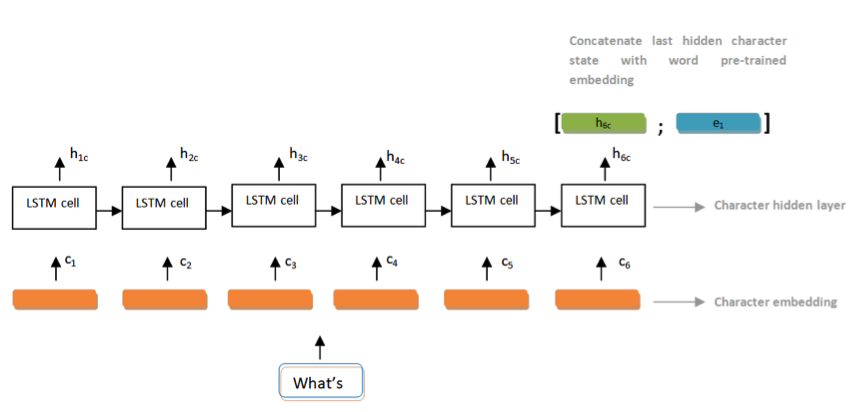}
	\caption{Character embeddings with LSTM}\label{fig:bilstm_crf_lstm}
\end{figure}

\noindent\textbf{Bi-LSTM--CRF with LSTM character embeddings}
As in the previous model we obtain a character level representation for each word;  character embedding 
is fed into an LSTM unit and the resulting vector $h_{6c}$ is the last hidden state of the character sequence \cite{lample2016neural}.
Then the resulting vector $h_{6c}$ is concatenated with the pre-trained word embedding as depicted in Figure~\ref{fig:bilstm_crf_lstm}.

\subsubsection{Intent Classification Model}
\begin{figure}[htb]
	\centering
	\includegraphics[width=\textwidth]{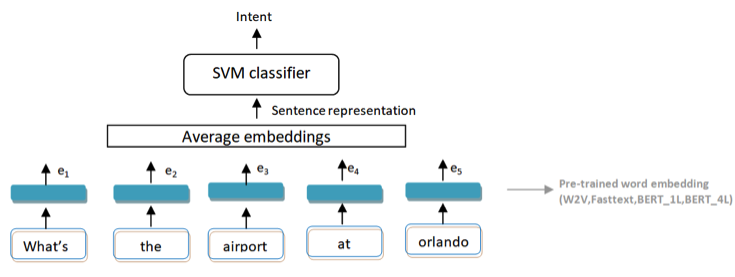}
	\caption{SVM intent classifier}\label{fig:svm_intent}
\end{figure}

For the task of IC we deploy an SVM classifier and compare the different word embeddings models at the 
initialization (Figure~\ref{fig:svm_intent}). Several approaches have been developed to solve multi-class problems through binary SVM 
techniques; in this paper we adopt the one-against-all approach \cite{hsu2002comparison}. 

In an $M$-class classification context this method processes $M$ binary problems: each one separates one class from 
the remaining $(M-1)$ ones. By training the corresponding $M$ SVM models the following decision functions can be obtained
\begin{align}
	&\bm{w}_{1}\:\, \cdot \Phi(\bm{x}) +b_{1} \notag \\
	&\bm{w}_{2}\:\, \cdot \Phi(\bm{x}) +b_{2}  \\
	&\:\, \vdots \notag \\
	&\bm{w}_{M}\cdot \Phi(\bm{x})+b_{M} \notag 
\end{align}
Finally, in the evaluation phase, a sample $\bm{x}$ is assigned to the class with the largest value of the decision function
\begin{align}
	\underset{i=1,\ldots,M}{\text{arg\,max}} \{\bm{w}_{i}\cdot \Phi(\bm{x})+b_{i} \}
\end{align}
We use the average of the word embeddings of the sentence to form its representation.

\subsubsection{Unified Model}
\begin{figure}[htb!]
	\centering
	\includegraphics[width=\textwidth]{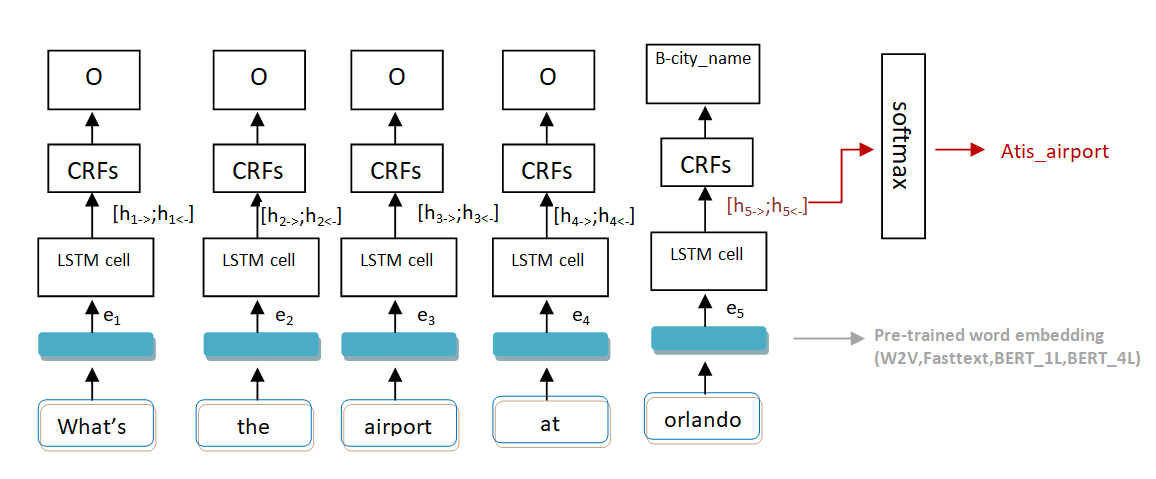}
	\caption{Unified model}\label{fig:model}
\end{figure}

The structure of our unified model is presented in Figure~\ref{fig:model}. In our model we employ an LSTM unit instead of a  
Gated Recurrent Unit (GRU) compared to previous approaches \cite{zhang2016joint}.
The input of the network is the text $S$ of an utterance, which is a sequence of words $w_{1},\ldots,w_{T}$, 
and $T$ is the length of the utterance. The network consists of two kinds of outputs, i.e. the predicted entity label 
sequence and predicted intent label. The bidirectional hidden states are shared by two tasks. On one hand, 
the hidden states capture the features at each time step, so they are directly used for predicting entities labels. 
On the other hand, we use the concatenated left and right last hidden state $h_{u} = [h_{\text{last}\to}; h_{\text{last}\leftarrow}]$ 
to acquire the representation of the whole sequence. The entity labels are predicted by the CRFs layer as in the previous models. 
For the intent prediction, the \emph{softmax} function is applied to the representation $h_{u}$ with linear transformation
 to give the probability distribution $y^{u}$ over the intent labels. Formally
\begin{align}
 	y^{u}=\text{softmax}(W^{u}h^{u}+b^{u})
\end{align}
where $W^{u}$ is transformation matrix  and  $b^{u}$ is bias vector.
 
The loss function of intents used is the categorical cross entropy. The categorical cross entropy loss measures the 
 dissimilarity between the true label distribution $y$ and the predicted label distribution $\hat{y}$.
 \begin{align}
 	L_{\text{cross-entropy}}(\bm{\hat{y}},\bm{y})=-\sum_{i} y_{i}log(\hat{y}_{i})
 \end{align}
 where $\bm{y} =(y_{1},\ldots,y_{n})$ is a vector representing the distribution over the labels $1,\ldots, n$ and 
 $\bm{\hat{y}} =(\hat{y}_{1},\ldots,\hat{y}_{n})$ is the classifiers output. 
 
The training target of the network is minimizing the sum of two losses i.e. $L_{\text{intent}}+L_{\text{entities}}$. 

\subsection{Transformer}

Transformer-based models and more specifically BERT based models for NER and IC implement the notion 
that the two tasks are strongly dependent of each other \cite{devlin2018bert}. BERT provides a powerful context-dependent 
sentence representation and can be used for various target tasks, i.e. NER and IC, through a fine-tuning procedure, 
similar to how it is used for other NLP tasks. 
We utilized two different BERT-based models to perform NER and IC on ATIS, for both English and Greek.

\subsubsection{BERT for Joint IC and NER}
The first model’s architecture largely implements the proposed pipeline in 
\emph{BERT for Joint Intent Classification and Slot Filling} \cite{chen2019bert}. 
The pre-trained BERT model provides a powerful context-dependent sentence 
representation and can easily be extended to a joint IC and NER model. 
The intent is predicted as
\begin{equation}
	y^{i}=\text{softmax} (W^{i}h_{1}+b^{i})
\end{equation}
where $h_{1}$ is the hidden state of the first special token (CLS). 

For NER, the final hidden states of other tokens $h_{2},\ldots,h_{T}$ are feeded into a softmax layer to classify over the labels
\begin{equation}
	y^{s}_{n}=\text{softmax}(W^{s}h_{n}+b^{s}), \quad n \in 1,\ldots, N
\end{equation}
where $h_{n}$ is the hidden state of the first subtoken of the word $x_{n}$. 

The goal is to maximize the conditional probability 
\begin{equation}
	p(y^{i},y^{s}\mid \bm{x})=p(y^{i}\mid \bm{x}) \prod_{n=1}^{N} p(y^{s}_{n}\mid\bm{x})
\end{equation}	 
The model is end-to-end  fine-tuned by minimizing the \emph{cross entropy loss}.
It is important to note that especially for the NER task, the labels' predictions are highly affected by the surrounding words which is why 
the addition of a CRF is explored aiming to improve the prediction of the entity labels on the top of the JointBERT model. For the 
token representation we used BERT embeddings for English \cite{devlin2018bert} and BERT embeddings 
for Greek \cite{koutsikakis2020greek}.

\subsubsection{Co-Interactive Transformer for Joint NER and IC}
The second model utilized for our experiments is a BERT-based model initially designed for Spoken Language Understanding (SLU) 
and manages to address the NER and IC tasks efficiently. 
The architecture of the Co-Interactive Transformer for Joint Intent Classification and Slot Filling incorporates a Bi-LSTM 
as a shared encoder \cite{qin2021co}. 
The Bi-LSTM consists of two LSTM layers. 
For an input sequence of $n$ tokens $\{x_{1}, x_{2},\ldots, x_{n}\}$ the Bi-LSTM reads it forwardly and backwardly to produce 
a series of context-sensitive  hidden states $H = \{h_{1},h_{2},\ldots,h_{n}\}$ by repeatedly applying the recurrence 
$h_{i} = \text{BiLSTM}(\phi^{\text{emb}}(x_{i})h_{i-1})$, where $\phi^{\text{emb}}$ represents the embedding function. 
Firstly, the model performs label attention over intent and entity labels to get explicit representations. 
Then, they are fed into the co-interactive layer to establish the dependency between the two. 
This produces the entity embedding matrix, $W^{S} \in R^{d\times | S^{\text{label}} |}$ and the intent embedding matrix  
$W^{I} \in R^{d\times | I^{\text{label}} |}$, where $d$ represents the hidden dimension and $S^{\text{label}}$ and $I^{\text{label}}$ 
represents the numbers of entity and intent labels, respectively.  

Intent label attention is calculated by the formula $A=\text{softmax}(HW^{v})$, where $H\in \mathbb{R}^{n\times d}$ is used 
as the query, $W^{v} \in \mathbb{R}^{d\times |v^{\text{label}}|} (v\in I or S)$ as the key and value in order 
to obtain the representation $H_{v}=H+A W^{v}$.

We can, therefore, compute $H_{I} \in \mathbb{R}^{n\times d}$ and  $H_{S} \in \mathbb{R}^{n\times d}$,
which are the explicit intent and entity representations and effectively capture semantic information. 
$H_{S}$ and $H_{I}$ are further used in the co-interactive attention layer in order to model the connection 
between entity and intent. 
Through the usage of linear projections, the matrices $H_{S}$ and $H_{I}$ are mapped to queries 
($Q_{S}$ and $Q_{I}$), keys ($K_{S}$ and $K_{I}$) and values ($V_{S}$ and $V_{I}$). The output of the model 
is a weighted sum of values
\begin{align}
	C_{S} &= \text{softmax} \left( \dfrac{Q_{S}K_{I}^{T}}{\sqrt{d_{k}}}\right) V_{I} \\
	H_{S}^{'} &= LN (H_{S}+C_{S}) 
\end{align}
which is the final entity representation. 

Here $LN$ represents the layer normalization function. By replacing $Q_{S}$ by $Q_{I}$,  $K_{S}$ by $K_{I}$ 
and $V_{S}$ by $V_{I}$ we obtain  the entity-aware intent representation. For English we used BERT 
and GloVe embeddings, 
and for Greek we used BERT 
and Word2vec embeddings 
to model the texts.


\subsection{Word embedding}

Word-embedding is an umbrella term for various language modelling and feature learning approaches that try to capture semantic 
and syntactic features of text through a vectorisation of text into vectors of real numbers (embeddings). 
The introduction and vast application of neural networks into NLP studies over the last decade with state-of-the-art performance in
various NLP tasks and their ability to capture refined characteristics and contextual cues  is substantially based on word-embedding. 
Several representation methods has been appeared in the literature where the most popular are Word2vec and its extensions 
Sentence2vec and Doc2vec, ELMo, BERT, fastText and GloVe, among others.

\subsubsection{Word2vec} 

Word2vec is probably the most popular word-embedding technique of text representation aiming to capture semantic content from 
text \cite{mikolov2013efficient,mikolov2013distributed}. It is an unsupervised approach that aims to capture the context 
of the words in a document employing the training of a shallow neural network with a large text corpus and providing a probability 
distribution for either (a) a word when the surrounding words are given (Continuous Bag-of-Words learning algorithm, CBOW) 
or (b) for a set of surrounding words when a single word is known (Continuous Skip-Gram algorithm, Skipgram). 
The desired embeddings are constructed in two steps: (a) Word2vec builds a vocabulary based on the words that appear in the 
corpus under investigation, and (b) either the CBOW or the Skip-gram algorithm are applied to learn the vector representations 
in a space with dimensions, usually up to $300$ \cite{mikolov2013efficient}. The model is trained with large textual corpora, e.g. 
all available Wikipedia articles in a certain language, in order to capture linguistic regularities that are of a global scope 
(within the given language). In advance, further training can be applied with the use of thematic textual collections  
to fine-tune the model for targeted domains of interest, e.g. technical documentation, reviews, etc.

\subsubsection{BERT}

Bidirectional Encoder Representation from Transformers (BERT) is a technique for training a general purpose 
language representation model for NLP \cite{devlin2018bert}. BERT elaborates pre-training contextual representations 
such as Semi-supervised Sequence Learning \cite{dai2015semi}, Generative Pre-Training \cite{radford2018improving},
ELMo \cite{peters2018deep} and ULMfit \cite{howard2018universal}. BERT is a deeply bidirectional, unsupervised learning 
representation pre-trained using only a plain text corpus instead of the vast amount of training data. 
It employs a Transformer architecture that allows massive parallelisation during training and produces global dependencies 
between input and output \cite{vaswani2017attention}. BERT has already presenting state-of-the-art results in major NLP tasks, 
such as question answering\footnote{\url{https://rajpurkar.github.io/SQuAD-explorer/}} 
and Natural Language Inference.\footnote{\url{https://www.nyu.edu/projects/bowman/multinli/}}

\subsubsection{fastText} 
The fastText library is a popular model used for very fast and efficient word embeddings 
and text classification \cite{joulin2017bag,bojanowski2017enriching}.
It assumes that a word is formed by $n$-grams, where $n$ ranges from $1$ to the length of word, 
and, thus, provides useful embeddings for previously unseen words, e.g. out of vocabulary words, 
languages with large vocabularies and many rare words, morphologically rich languages or typos. 
The fact that combines high efficiency in terms of accuracy in 
many order of magnitude faster for training and evaluation compared to other standard deep learning classifiers, 
the availability of pre-trained models for almost $300$ different languages and its performance on small datasets has made it
popular for various NLP applications.  

\subsubsection{Glove}

GloVe, coined from global vectors, is a model for distributed word representations \cite{pennington2014glove}. 
The model is an unsupervised learning algorithm for obtaining vector representations for words. 
Training is performed on aggregated word-word co-occurrence statistics from a corpus, and the resulting representations showcase interesting linear substructures of the word vector space. 
Essentially, it is a log-bilinear model with a weighted least-squares objective. 
The main intuition underlying the model is the simple observation that ratios of word-word co-occurrence probabilities have the potential for encoding some form of meaning. 
GloVe can be used to find relations like synonyms, cities, company-product relations, etc. 
GloVe has proved to yield promising results and has been praised for its speed. 
For our experiments we harvested the 300-dimensional version of the GloVe embedding vectors. 


\subsection{Preprocessing and word embeddings initialization}

The standard ATIS dataset after the removal of duplicates results in 5473 sentences (all words were lowercased). 
The test set was the $30\%$ of the corpus. For the Word2vec initialization we used the Google News Word2vec 
pre-trained model.\footnote{\url{https://drive.google.com/file/d/0B7XkCwpI5KDYNlNUTTlSS21pQmM/edit?usp=sharing}} 
For the fastText initialization we use a model which is pre-trained in Common Crawl and Wikipedia.\footnote{\url{https://dl.fbaipublicfiles.com/fasttext/vectors-crawl/cc.en.300.vec.gz}}

For the translation of ATIS in Greek language, all words were lowercased and the accents removed. 
For the Word2vec initialization we trained a Word2vec model in the Greek Wikipedia. 
For the fastText initialization we used the model in Common Crawl 
and Wikipedia.\footnote{\url{https://dl.fbaipublicfiles.com/fasttext/vectors-crawl/cc.el.300.vec.gz}} 
For the GloVe initialization we used the 300 dimensional vectors from the model in Wikipedia 2014 and Gigaword 5.\footnote{\url{https://nlp.stanford.edu/data/glove.6B.zip}}
Regarding the BERT initialization we used the Google pre-trained BERT$_\text{BASE}$  model in Wikipedias of 104 different languages
to extract the token embeddings of both corpuses 
(English and Greek)\footnote{\url{https://storage.googleapis.com/bert_models/2018_11_23/multi_cased_L-12_H-768_A-12.zip}} with the exception of the JointBERT Transformer model, where we opted to use the GreekBERT vectors, which were produced from pretraining on a corpora that included the Greek part of Wikipedia, the Greek part of European Parliament Proceedings Parallel Corpus and the Greek part of OSCAR, a cleansed version of Common Crawl.\footnote{\url{https://huggingface.co/nlpaueb/bert-base-greek-uncased-v1}}
Here we have to note that English Wikipedia includes 6,012,532 articles in addition to Greek Wikipedia which includes 
173,380 articles.\footnote{\url{https://meta.wikimedia.org/wiki/List_of_Wikipedias}}  
For our LSTM experiments, two approaches were used. In the first approach we averaged the token vectors from the last 4 layers of BERT (which is denoted as BERT 4L in our tables) and in the second 
approach we used only the token vector from the last layer (which is denoted as BERT 1L in our tables). 
Then, in order to produce the word vector, we averaged the tokens vectors of each word.


\section{Experimental Results}\label{sec:expres}

\subsection{Experimental setup}

Experiments were conducted mainly on Google Colab.\footnote{\url{https://colab.research.google.com}} 
The GPUs available often include Nvidia\copyright \ K80s, T4s, P4s and P100s 
(the user is not allowed to choose the type of the GPU at any given time). The RAM available for our experiments was 12 Gb and the CPU frequency was 2.30 GHz (Intel\copyright \ Haswell family).

For Bi-LSTM experiments the maximum word sequence length were 
set to $35$ while the maximum character sequence were set to $30$. Word embedding and character dimensions are 
presented in Tables~\ref{tab:wedim}~and~\ref{tab:chdim}.

\subsection{Datasets}

We perform the experiments on the Airline Travel Information System corpus (ATIS) a popular, benchmark dataset in NER research area \cite{price1990evaluation}. The domain of the dataset is airline travel and the intents include finding a flight, inquiring about airlines or their services, asking about ticket prices, etc. The ATIS pilot corpus is a corpus designed to measure progress in Spoken Language Systems that include both a speech and a natural language component. Due to the nature of the dataset, it has been used extensively to train, test and develop NER and IC networks and question answering systems, for both spoken and written language. As the dataset consists of audio recordings and their corresponding manual transcripts and every utterance is completely labeled, it is convenient for both types of approaches. For the needs of our experiments, we harvested the textual data of ATIS and we translated it to Greek employing a panel of 5 persons in order to optimize the quality of the translation. 

An example of ATIS' sentence in both English and Greek languages is presented in Table~\ref{tab:IOB}  along with the appropriate In/Out/Begin (IOB) representation. It is important to note here that we didn't use the public version of the ATIS 
dataset which comprises of 6763 sentences but a modified version where 1290 duplicate utterances have been removed resulting in a "clean" version which consists of 5473 unique sentences. This processed version has been translated in Greek language (5473 unique sentences) and used in the respective experiments.  Both datasets are available from the corresponding author on reasonable request or directly through the following hyperlink.\footnote{\url{https://msensis.com/research-and-development/downloads}}

\begin{table}
\centering
\begin{adjustbox}{max width=\textwidth}
\begin{tabular}{lc||lc}
	\hline\hline
	Intent & \textsf{atis\_flight} & 	Intent & \textsf{atis\_flight} \\
	\hline
	Sentence (EN) & Named Entity & Sentence (GR) & Named Entity \\
	\hline\hline
	Show & O & \textgreek{Δείξτε} & O \\
	Sunday & \textsf{B-depart\_date.day\_name} & \textgreek{τις} & O \\
	flights & O & \textgreek{πτήσεις} & O \\
	from & O & \textgreek{της} & O \\
	Seattle & \textsf{B-fromloc.city\_name} & \textgreek{Κυριακής} & \textsf{B-depart\_date.day\_name}\\
	to & O & \textgreek{από} & O  \\
	Chicago & \textsf{B-toloc.city\_name} & \textgreek{Σιατλ} & \textsf{B-fromloc.city\_name} \\
	& & \textgreek{για} & O \\
	& & \textgreek{Σικάγο} & \textsf{B-toloc.city\_name} \\
	\hline\hline 
\end{tabular}
\end{adjustbox}	
	\caption{In/Out/Begin (IOB) representation for an example sentence in ATIS-EN and ATIS-GR respectively  }\label{tab:IOB}
\end{table}


\subsection{Evaluation metrics}

The performance of the intents and entities recognition models under inspection is assessed with the employment of the 
\emph{accuracy} metric defined as
\begin{align}
\text{accuracy} = \dfrac{TP+TN}{TP+FP+TN+FN} 
\end{align}
Also we calculate \emph{precision}, \emph{recall} and we present their harmonic mean, the balanced $F$-\emph{score}
defined as
\begin{align}
	F\text{- score} = 2\times \dfrac{\text{precision}\times \text{recall}}{\text{precision} + \text{recall}}
\end{align}
where 
\begin{align*}
	\text{precision} &=\dfrac{TP}{TP+FP} \\
	\text{recall} &=\dfrac{TP}{TP+FN} 
\end{align*}
and $TP$, $FP$ are the true and false positives, while $TN$, $FN$ are the true and false negatives, respectively. 

There are some utterances in ATIS dataset that have more than one intent labels. An utterance is counted a correct in the 
classification task if any ground truth label is predicted correctly. A slot is considered to be correct if its range and type are 
correct.

\subsection{Training Hyperparameters and Evaluation Tools}

The hyperparameters for LSTM and CNN implementations are presented in Tables~\ref{tab:lstmdim}~and~\ref{tab:cnndim}. Table~\ref{tab:transdim} shows the hyperparameters for the Transformer-based networks. All hyperparameters were sampled from predefined suitable value ranges. In some experiments they were further fine-tuned to optimize performance. Both families of networks were evaluated with the \emph{scikit-learn} library, which currently is the most widely used tool for measuring model performance.

\begin{table}[h!]
    \centering
\begin{minipage}[t]{0.48\linewidth}\centering
\begin{tabular}{ l c }
\hline
	Method  &  Embeddings dimensions \\
\hline
	Word2vec &  $300$ \\
	fastText & $300$      \\
	BERT  & $768$    \\
\hline
\end{tabular}
	\caption{Word embeddings}\label{tab:wedim}
\end{minipage}\hfill
\begin{minipage}[t]{0.48\linewidth}\centering
\begin{tabular}{ l c }
\hline
	Method    & Embeddings dimensions \\
\hline
	LSTM &  $20$ \\
	CNN & $10$    \\
\hline
\end{tabular}
	\caption{Character embeddings }\label{tab:chdim}
\end{minipage}
\end{table}

\begin{table}[h!]
    \centering
\begin{adjustbox}{max width=\textwidth}    
\begin{tabular}{ l c c c}
\hline
	Model  &  Learning rate & Optimizer & LSTM units \\
\hline
	BiLSTM\hfill &  $0.001$ & rms-prop & $100$ \\
	BiLSTM with LSTM char embeddings & $0.001$ & adam & \phantom{0}$50$      \\
	BiLSTM with CNN char embeddings & $0.001$ & adam & \phantom{0}$50$ \\
	Unified model & $0.001$ & adam & $100$ \\
\hline
\end{tabular}
\end{adjustbox}
	\caption{LSTM parameters}\label{tab:lstmdim}
\end{table}

\begin{table}[h!]
	\centering
\begin{tabular}{ l c }
\hline
	filters    &  $10$ \\
	stride &  \phantom{0}$1$ \\
	kernel size & \phantom{0}$3$    \\
	dropout & \phantom{00.}$0.5$	\\
\hline
\end{tabular}
	\caption{CNN parameters}\label{tab:cnndim}
\end{table}

\subsection{Bi-LSTM Results}

The results for the NER models are demonstrated in Table~\ref{tab:res1}. These include the 
Bi-LSTM--CRF model with and without character embeddings obtained from CNN and LSTM, and the employment 
of three different language models, i.e. Word2vec, fastText and BERT. 

\begin{table}
\begin{small}
\centering 
\begin{adjustbox}{max width=\textwidth}
\begin{tabular}{l|cc|cc|cc|cc}
	\hline\hline
	& \multicolumn{2}{c}{Word2vec} & \multicolumn{2}{c}{fastText} & \multicolumn{2}{c}{BERT (1L)} & \multicolumn{2}{c}{BERT (4L)}\\
	\hline
	 & Prec & $F$-score & Prec  & $F$-score & Prec  & $F$-score & Prec  & $F$-score \\
	\hline\hline
	BiLSTM--CRF & \multicolumn{8}{c}{ATIS-EN} \\
	\hline\hline
	Character embeddings & & & & & & & & \\ 
	No  & 0.889 & 0.899 & 0.945 & 0.947 & 0.968 & 0.968 & 0.972  & 0.972 \\
	CNN  & 0.895 & 0.905 & 0.932  & 0.939	& 0.909  & 0.916 & 0.913 	& 0.921 \\
	LSTM  & 0.867  & 0.884 & 0.916  & 0.921  & 0.964	& 0.956 & 0.962 	& 0.964 \\	
	\hline\hline
	BiLSTM--CRF & \multicolumn{8}{c}{ATIS-GR} \\
	\hline\hline
	Character embeddings & & & & & & & & \\ 
	No  & 0.901 & 0.910 & 0.891  & 0.895 & 0.908  & 0.913 & 0.908 	& 0.911 \\
	CNN  & 0.956  & 0.958 & 0.934 	& 0.943 & 0.949  & 0.955 & 0.955 & 0.959 \\
	LSTM  & 0.944  & 0.948	& 0.922  & 0.928 & 0.936 & 0.940 & 0.953  & 0.957 \\
	\hline\hline	
\end{tabular}
\end{adjustbox}
\end{small}
	\caption{NER models (ATIS-EN and ATIS-GR)}\label{tab:res1}
\end{table}

\begin{table}
\centering 
	\begin{small}
\begin{adjustbox}{max width=\textwidth}
	\begin{tabular}{cccccccc}
	\hline
	\multicolumn{8}{c}{ATIS-EN}\\	
	\hline
	\multicolumn{2}{c}{Word2vec} & \multicolumn{2}{c}{fastText}  & \multicolumn{2}{c}{BERT (1L)}   & \multicolumn{2}{c}{BERT (4L)}\\
	\hline
	Precision & $F$-score & Precision & $F$-score & Precision & $F$-score & Precision & $F$-score \\
	\hline 
	0.95	& 0.94 & 0.96 & 0.96 & 0.96 & 0.96 & 0.95 & 0.95 \\
	\hline
	\multicolumn{8}{c}{ATIS-GR}\\	
	\hline
	\multicolumn{2}{c}{Word2vec} & \multicolumn{2}{c}{fastText}  & \multicolumn{2}{c}{BERT (1L)}   & \multicolumn{2}{c}{BERT (4L)}\\
	\hline
	Precision & $F$-score & Precision & $F$-score & Precision & $F$-score & Precision & $F$-score \\
	\hline 
	0.93	& 0.93 & 0.96 & 0.95 & 0.95 & 0.95 & 0.95 & 0.95 \\
	\hline	
	\end{tabular}
\end{adjustbox}
	\end{small}
		\caption{IC with SVM (ATIS-EN and ATIS-GR)}\label{tab:res2}
\end{table}

\begin{table}
	\centering
	\begin{small}
\begin{adjustbox}{max width=\textwidth}
	\begin{tabular}{l|cccccccc}
	\hline
	\multicolumn{9}{c}{ATIS-EN}\\	
	\hline	
	&	\multicolumn{2}{c}{Word2vec} & \multicolumn{2}{c}{fastText}  & \multicolumn{2}{c}{BERT (1L)}   & \multicolumn{2}{c}{BERT (4L)}\\
	& 	Precision & $F$-score & Precision & $F$-score & Precision & $F$-score & Precision & $F$-score \\
	Entities & 0.918 & 0.922 & 0.959 & 0.960 & 0.969 & 0.971 & 0.974 & 0.975 \\
	Intents & 0.896 & 0.905 & 0.934 & 0.936 & 0.952 & 0.950 & 0.955 & 0.953 \\
	Average & 0.907 & 0.913 & 0.946 & 0.948 & 0.960 & 0.960 & 0.964 & 0.964 \\
	\hline
	\multicolumn{9}{c}{ATIS-GR}\\	
	\hline	
	&	\multicolumn{2}{c}{Word2vec} & \multicolumn{2}{c}{fastText}  & \multicolumn{2}{c}{BERT (1L)}   & \multicolumn{2}{c}{BERT (4L)}\\
	& 	Precision & $F$-score & Precision & $F$-score & Precision & $F$-score & Precision & $F$-score \\
	Entities & 0.960 & 0.961 & 0.954 & 0.954 & 0.953 & 0.955 & 0.958 & 0.960 \\
	Intents & 0.930 & 0.934 & 0.933 & 0.938 & 0.918 & 0.921 & 0.926 & 0.931 \\
	Average & 0.945 & 0.947 & 0.943 & 0.946 & 0.935 & 0.938 & 0.942 & 0.945 \\	
	\hline
	\end{tabular} 
\end{adjustbox}	
        \end{small}
        \caption{Unified model}\label{tab:res3}
\end{table}        

The NER model with BERT (4L) initialization outperforms all other models in both languages in entities recognition. 
For intents classification, as can be observed in Table~\ref{tab:res2}, there are no significant differences. 

The unified model with the BERT (4L) initialization outperforms all other approaches in the case of the English language (ATIS EN). 
In the case of the Greek language there is no superior architecture. For entities and average (both entities and intents) the 
Word2vec initialization provides the best results; just for intents the fastText initialization performs slightly better. 
The metrics yielded by the unified model are presented in Table~\ref{tab:res3} for both languages.

Comparing the results between the two datasets, we notice that entity and intent scores are slightly better in English language experiments compared with Greek language ones (except all Bi-LSTMs--CRF models (Word2vec initialization) and the unified model (Word2vec initialization)).

\subsection{Transformer Results}
The results yielded by the Transformer-based models are demonstrated in the tables below. Table~\ref{tab:transen} presents the experimental results we obtained for the English version of ATIS and Table~\ref{tab:transgr} for the Greek version of ATIS. The various word-level representations are showcased.

Overall accuracy refers to the complete prediction of the given sentence.
A sentence is considered correct if the network manages
to predict both the intent as well as every entity found in the utterance.
Even a single mislabelled word is enough to deem the prediction of the whole utterance as false,
which is often an impractical approach when building joint NER and IC systems, as the occasional
minor error is expected and can be handled easily.
Here we opt to present the overall accuracy results even though that as expected are relatively low compared to the other metrics

As expected, the Transformer family of networks yields better results in comparison to the LSTM family, and JointBERT in particular appears to produce the best results for both languages regarding almost every metric we examined. Regarding English specifically, it is interesting to note that overall accuracy is significantly higher in comparison to the Co-Interactive model. Lastly, concerning the Co-Interactive model, GloVe embeddings yielded slightly better results compared to BERT embeddings for the English version, unlike the Greek version of ATIS, where BERT embeddings had slightly higher metrics compared to Word2vec embeddings. 
\\
\begin{table}
\begin{small}	
	\centering
\begin{adjustbox}{max width=\textwidth}
	\begin{tabular}{lccc}
	\hline
	Model & Entities ($F_{1}$/Prec) & Intents (Acc) & Overall (Acc) \\
	\hline
	Co-Interactive Transformer (GloVe embeddings)  & 0.989/0.989 	& 0.977 	& 0.909 \\
	Co-Interactive Transformer (BERT embeddings)  & 0.987/0.988 	& 0.975 	& 0.905 \\
	JointBERT (BERT embeddings) 	& 0.993/0.993 	& 0.985 	& 0.937 \\ 	 
	JointBERT (BERT embeddings and CRF) 	& 0.993/0.993 	& 0.981 	& 0.936 \\	  	
	\hline 
	\end{tabular}
\end{adjustbox}
	\caption{ATIS-EN}\label{tab:transen}
	\end{small}
\end{table}

\begin{table}
\begin{small}	
	\centering
\begin{adjustbox}{max width=\textwidth}
	\begin{tabular}{lccc}
	\hline
	Model & Entities ($F_{1}$/Prec) & Intents (Acc) & Overall (Acc) \\
	\hline
	Co-Interactive Transformer (Word2vec embeddings)  & 0.980/0.980 	& 0.968 	& 0.848 \\
	Co-Interactive Transformer (BERT embeddings)  & 0.984/0.984 	& 0.979 	& 0.881 \\
	JointBERT (GreekBERT embeddings) 	& 0.988/0.988 	& 0.975 	& 0.888 \\ 	 
	JointBERT (GreekBERT embeddings and CRF) 	& 0.989/0.989 	& 0.973 	& 0.896 \\	 	
	\hline 
	\end{tabular}
\end{adjustbox}
	\caption{ATIS-GR}\label{tab:transgr}
	\end{small}
\end{table}	

The  performance of the Transformer models is higher, for both entities and intents, in the case of ATIS-EN compared with experiments performed in ATIS-GR. The differences in the overall accuracy scores are augmented compared to the differences in the intents and entities scores due to the nature of the overall accuracy definition. 

The GreekBERT word embeddings used for the Greek version of ATIS for the JointBERT model specifically were harvested 
from AUEB \cite{koutsikakis2020greek}, which is part of Hugging Face's Transformers repository.\footnote{\url{https://huggingface.co/nlpaueb/bert-base-greek-uncased-v1}} 

\begin{table}[h!]
    \centering
\begin{tabular}{ l c c c}
\hline
	Model  &  Learning rate & Optimizer  \\
\hline
	JointBERT\hfill &  $0.00005$ & adam \\
	Co-Interactive Transformer & $0.001$\phantom{00} & adam     \\
\hline
\end{tabular}
	\caption{Transformer parameters}\label{tab:transdim}
\end{table}


\section{Conclusions}\label{sec:conc}

The main goal of this work is to present a comparative study between state-of-the-art  deep learning models for NER and IC tasks. 
Also to evaluate the effectiveness of the proposed methods in two different languages, i.e. English which is a typical low inflection language versus Greek which is a typical high inflection language.  
In our experiments with Bi-LSTM networks it was the BERT embeddings that provided the best results in both languages under inspection. However, character embeddings (CNN and LSTM) improved models' performance only in the case of the Greek language dataset. 
Experiments with Transformer-based models also display impressive results and significant improvement for English compared to 
Bi-LSTM models. 
In our knowledge there is no other benchmark dataset available (for NER and IC) in both languages. 

It is evident that deep learning methods are becoming very popular in the fields of NER and IC, gradually replacing rule-based 
approaches based on domain-specific gazetteers and syntactic-lexical patterns. 
There are three core strengths in deep learning that establish its dominance over the previous techniques. 
Firstly, NER benefits from non-linear transformation which enable the learning of complex and intricate features about the data by implementing non-linear activation functions. 
Secondly, deep learning saves significant amounts of time and effort, as feature-based approaches require considerable 
domain expertise and engineering skill. 
Thirdly, deep neural NER models can be trained in an end-to-end paradigm, by gradient descent, effectively enabling 
the design of complex NER systems. 
The incorporation of word- 
and character-level representations in the form of embedding vectors 
has been vital for the improvement of NER and IC. 
Our experiments showcase that both families of Bi-LSTM and Transformer networks are suitable for NER and IC tasks 
and relative business bilingual applications since they capture the complex structure of natural language texts in different languages.


Further investigation of JointBERT model to explore the dependency between intent and entities via label attention is part of future work. 
Morover, it is worth exploring the combination of different word vectorization such as ELMo and OpenAI \cite{neelakantan2022text}.





\section*{Acknowledgements}
\label{sec:acknow}
This research has been co‐financed by the European Regional Development Fund of the European Union and Greek national funds through the Operational Program Competitiveness, Entrepreneurship and Innovation, under the call RESEARCH -- CREATE -- INNOVATE (project code: T1EDK-05732).

\bibliographystyle{abbrv}
\bibliography{mybib}

\end{document}